\pdfoutput=1

\documentclass[11pt]{article}

\usepackage{ACL2023}

\usepackage{times}
\usepackage{float}
\usepackage{xcolor}
\usepackage{latexsym}
\usepackage{amssymb}
\usepackage{amsmath}
\usepackage{amsfonts}
\usepackage{dsfont}
\usepackage{graphicx}
\usepackage{caption}
\usepackage{subcaption}
\usepackage{multirow}
\usepackage{booktabs}
\usepackage{algorithm}
\usepackage{algpseudocode}
\usepackage{tabularx}
\usepackage{hyperref}
\usepackage{pifont}
\usepackage{natbib}
\DeclareMathOperator*{\argmax}{arg\,max}

\usepackage{siunitx} 

\definecolor{lightgray}{gray}{0.9}
\usepackage[T1]{fontenc}

\usepackage[utf8]{inputenc}

\usepackage{microtype}

\usepackage{inconsolata}
\usepackage{lipsum}

\title{Towards Better Question Generation in QA-based Event Extraction}

\author{Zijin Hong \\
  Jinan University \\
  \texttt{hongzijin@stu2020.jnu.edu.cn} \\\And
  Jian Liu\footnotemark[2] \\
  Beijing Jiaotong University \\
  \texttt{jianliu@bjtu.edu.cn} \\}

\begin{document}
\maketitle
\renewcommand{\thefootnote}{\fnsymbol{footnote}} 
\footnotetext[2]{Corresponding author}
\renewcommand{\thefootnote}{\arabic{footnote}} 

\begin{abstract}
Event Extraction (EE) is an essential information extraction task that aims to extract event-related information from unstructured texts.
The paradigm of this task has shifted from conventional classification-based methods to more contemporary question-answering-based (QA-based) approaches. 
However, in QA-based EE, the quality of the questions dramatically affects the extraction accuracy, and how to generate high-quality questions for QA-based EE remains a challenge.
In this work, to tackle this challenge, we suggest four criteria to evaluate the quality of a question and propose a reinforcement learning method, \textbf{RLQG}, for QA-based EE that can generate generalizable, high-quality, and context-dependent questions and provides clear guidance to QA models.
The extensive experiments conducted on ACE and RAMS datasets have strongly validated our approach's effectiveness, which also demonstrates its robustness in scenarios with limited training data. The corresponding code of RLQG is released for further research\footnote{\href{https://github.com/Rcrossmeister/RLQG}{https://github.com/Rcrossmeister/RLQG}}.

\end{abstract}

\section{Introduction}
\begin{figure}[t]
    \centering
    \includegraphics[width=0.98\linewidth]{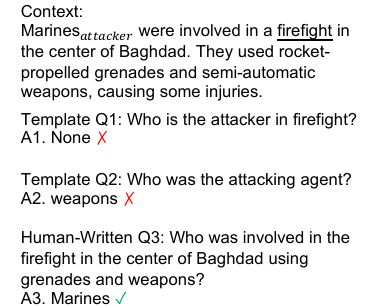}
    \caption{An EE example from ACE dataset, with ``firefight'' being the event trigger and ``Marines'' being the event argument fulfilling the \textit{Attacker} role.}
    \label{fig:intro}
\end{figure}

Event extraction (EE), an essential information extraction task, aims to extract event-related information (often called event arguments) from a given context.
Recently, the paradigm of EE has shifted from conventional classification-based approaches~\cite{li-etal-2013-joint, chen-etal-2015-event, nguyen-etal-2016-joint-event} to question-answering-based (QA-based) methods~\cite{du-cardie-2020-event, liu-etal-2020-event, liu-etal-2021-machine, ma-etal-2022-prompt, lu-etal-2023-event}.
For example, given a sentence: ``\textit{Marines were involved in a firefight in the center of Baghdad}'', to extract the role \textit{Attacker}, the method first generates the question ``\textit{Who is the attacker in firefight?}'' and then uses a question-answering (QA) model to obtain an answer as the corresponding result (as shown in Figure~\ref{fig:intro}).
This paradigm has demonstrated great success in various scenarios/domains.

Nevertheless, one of the biggest challenges in QA-based EE is obtaining ``good questions'' to guide the QA model, and the quality of the questions often significantly affects the results.
Previous methods typically use well-designed templates to generate such questions~\cite{du-cardie-2020-event, zhou-etal-2022-large, liu-etal-2023-prompt}, which, however, often lead to rigid and less context-dependent questions.
For example, Figure~\ref{fig:intro} gives two questions, Q1 and Q2, based on templates, leading to incorrect answers.
In our pilot experiments, we show that template-based questions lead to about 60\% errors even using a powerful proprietary QA model like GPT-4~\cite{openai-2023-gpt}, as shown in Figure~\ref{fig:QA-ablation}.

In this paper, we explore effective methods towards generating better questions for QA-based EE. 
First, we propose four criteria for a good question:
1) Fluency: A question should be fluent in natural languages to be well addressed by a QA model.
2) Generalizability: Methods for question generation should apply to unseen contexts and roles beyond those in training~\cite{du-cardie-2020-event}.
3) Context dependence: Questions should be consistent with the context and include necessary information to be correctly addressed by a QA model.
4) Indicative guidance for QA models: Questions should offer clear guidance for QA models to find answers~\cite{kojima-etal-2022-zeroshots}.
We then seek to build a model towards satisfying each aspect.

Methodologically, we develop a sequence-to-sequence-based text generation model that can learn from template questions, allowing it to generate more fluent questions and generalize to previously unexplored roles. 
Nonetheless, more is required to produce high-quality questions.
Further, to meet the criteria of context dependence and indicative guidance, we developed a \textbf{R}einforcement \textbf{L}earning-based \textbf{Q}uestion \textbf{G}eneration framework, \textbf{RLQG} to refine the generation process~\cite{christiano-etal-2017-rlhf,rafailov-etal-2023-direct}.
Specifically, an inverse prompting mechanism is proposed to evaluate whether the question matches the context, and a question-answering reward is used to quantify the degree to which the question is indicative.
We choose a positive and negative question pair based on the above two mechanisms and utilize these as signals to fine-tune the model, biasing it toward generating context-dependent and suggestive questions providing indicative guidance.

Finally, the effectiveness of our method has been verified on two widely used EE benchmarks.
According to the results, on the full ACE~\cite{doddington-etal-2004-automatic} and RAMS~\cite{ebner-etal-2020-multi} benchmarks, our method outperforms previous methods by 2.69\% and 1.96\%.
More importantly, we show that our method is particularly effective in data-scarce scenarios -- with only 40\% of the training data, we achieved the same performance as previous works.
Additionally, we show that we can achieve good performance based on simple questions without excessive manual intervention.

In summary, the contributions of our work are three-fold:
\begin{itemize}
    \item We revisit question generation for QA-based EE and suggest four question evaluation criteria. We design a model that can generate better questions with these as guidance. 
    
    \item We introduce a reinforcement learning framework for better question generation for EE, which is considered context-dependent and indicative of question generation.
    
    \item We have verified the effectiveness of our method on different benchmarks, and show its capability to handle the more challenging data-scarce scenario.

\end{itemize}
\section{Related Work}

\subsection{QA-Based Event Extraction}

Event extraction is an information extraction task focusing on extraction, particularly event information. Traditionally, methods formulate it as a classification problem~\cite{ahn-2006-stages, li-etal-2013-joint, chen-etal-2015-event, nguyen-etal-2016-joint-event}, but recent methods have started a QA-based paradigm.
The core is to generate a question to find each argument.
For example, \cite{liu-etal-2020-event, lyu-etal-2021-zero} convert EE tasks into machine reading comprehension using simple questions that are highly generic. Then~\cite{li-etal-2020-event} reformulate the task as multi-turn question answering, finish the trigger identification, and argument extraction by asking different questions. Recently, \cite{du-cardie-2020-event} and~\cite{lu-etal-2023-event} studied the effect on question quality in question answering; they trained a question generation (QG) model to generate a better question and also fine-tuned a model to do the question answering to finish the EE task. 
Despite the above advances, there still exists a challenge regarding how to generate a ``good'' question and even what the definitions of ``good questions'' are.
In this work, we provide four criteria for question generation and build a model towards satisfying each aspect.

\begin{figure*}[!htbp]
    \centering
    \includegraphics[width=1\linewidth]{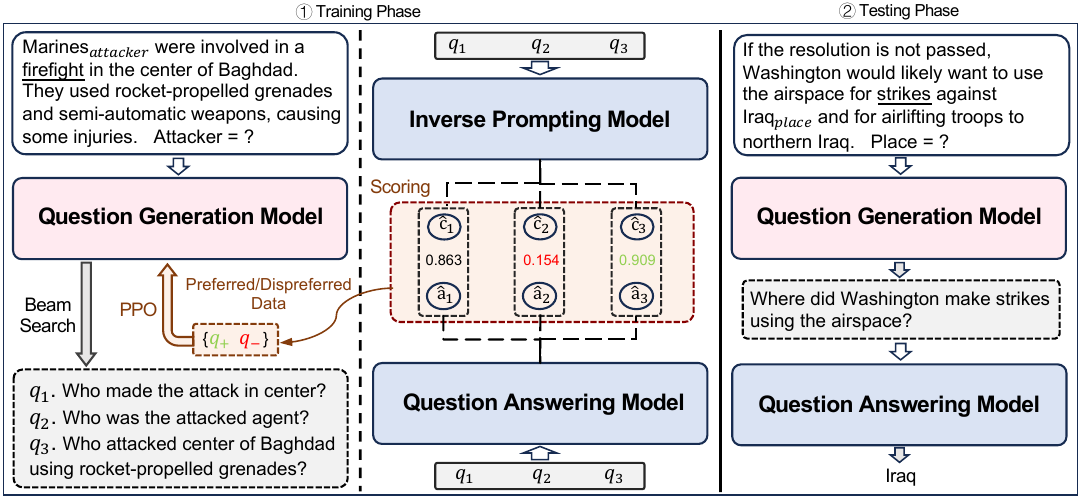}
    \caption{The overview of our proposed method, comprises: 1) Training phase, including supervised fine-tuning and reinforcement learning refining for a question generation model. 2) Testing phase, a final (off-the-shelf) question-answering model predicts the final answer based on the given context and question.}
    \label{fig:framework}
\end{figure*}

\subsection{Prompt Engineering for LLMs}
Our work also relates to prompt engineering in LLMs. 
The prompt is the natural language input for the language model. Previous research has proven the efficiency of prompt engineering~\cite{radford-etal-2019-unsupervisedmultitask,liu-etal-2023-prompt}; practical prompt engineering can drastically improve the efficiency and output quality of language models, making them more useful across a wide range of applications, from creative writing and content generation~\cite{zou-etal-2021-inverseprompting} to technical problem-solving and data analysis \cite{chen-etal-2023-program}. Recently, large language models (LLMs) became a main research object in language model study; as a chat model with general capacity in NLP task, prompt study began a crucial challenge in improving the performance of LLM's response~\cite{wei-etal-2022-zeroshotlearners}. Studies like Chain of Thoughts~\cite{wei-etal-2022-cot} and Retrieval Augmented Generation~\cite{lewis-etal-2020-rag} focus on involving information that helps the LLM for better contextual information for natural language reasoning and understanding. 
Our work is a particular case for generating prompts in EE to trigger LLM for information extraction.
\section{Proposed Method}

The overview of our method is visualized in Figure~\ref{fig:framework}, which contains three main modules:
\begin{itemize}
    \item A supervised fine-tuned (SFT) question generation model converts a given role and corresponding context into a question.

    \item A reinforcement learning (RL) framework refines the question generation through inverse prompting and question-answering rewards.

    \item A final (off-the-shelf) question-answering model to generate the final answer based on the given context and question.
\end{itemize} 
We detail each part in the following.

\subsection{Question Generation Module}\label{sec:qg}

The question generation module aims to generate a question for a semantic role.
Assuming the context is $c$, annotated with an event trigger $t$, our goal is to generate a question $q$ for the particular role $r$.

\paragraph{SFT over Template Questions.}
We utilize a general sequence-to-sequence framework as the backbone of the question generation model, and to make a good starting point, we use template questions as the targets for learning.
Particularly, for a context, trigger, role triplet ($c'$, $t'$, $r'$) in the training set, we construct the following input:
\begin{equation}\label{eq:prompt}
    p_{(c', t', r')} = \text{role}: []_{r'} \oplus \text{trigger}: []_{t'} \oplus \text{context}: []_{c'}
\end{equation}
Where $[]_{c'/t'/r'}$ denotes a slot to fill the context/trigger/role respectively, $\oplus$ is a concatenation operator. 
We then adopt the following template question\footnote{In practice, we can apply more complex templates~\cite{lu-etal-2023-event} as shown in Section~\ref{sec:exp setup}.} as the target:
\begin{center}
   $q'_{(r', t')} = $ WH\footnote{Interrogative pronoun.} is the $[]_{r'}$ in the $[]_{t'}$ event?
\end{center}
The question generation model learns a map from $p$ to $q'$ and is expressed as a probabilistic model. At testing time, the generated question $q$ is formulated as follows:
\begin{equation}\label{eq:qg}
    q = \argmax_{\tilde{q}} \operatorname{Pr}(\tilde{q} \mid p\;; \theta),
\end{equation}
where $\theta$ denotes the parameter set of the model.
In practice, we utilize LLaMA-2~\cite{touvron-etal-2023-llama2} as the base model, and the learning is performed by minimizing the cross-entropy loss function $\mathcal{L}_{c}$ over each training instance:
\begin{equation}
\begin{aligned}
    \mathcal{L}_{c} &= -\sum_{(c', t', r') \in D}\log \operatorname{Pr}(q'_{(r', t')} \mid p_{(c', t', r')}\;; \theta),
\end{aligned}
\end{equation}
where $\operatorname{Pr}(q \mid p)$ denotes the probability of generating $q$ by given $p$, $D$ is the training set.

\paragraph{Beam Search Augmentation.}\label{sec:BSG}
Considering that the standard question generation method can only get one question with the highest probability, to increase diversity, we adopt beam search augmentation to generate multiple sentences for further use.
Remarkably, at each step of beam search, it keeps track of the $N_{beam}$ most promising sequences, and $N_{beam}$ is the beam size. 
And therefore, for each ($c'$, $t'$, $r'$), it can generate a set of candidate questions $\mathcal{Q}$ = $\{q_1, q_2, ..., q_n\}$, $n$ is the number of return questions given $p$.

\subsection{RL for QG Refinement}\label{sec:RL}
To refine the questions, we build an RL framework with inverse prompting and question-answering reward.
Our motivation is that if a question is context-dependent and indicative enough, we can use it to recover the context to some extent. When used as input to the QA model, it can yield the correct answer (during training).
Then, we integrate the RL framework with the above two rewarding mechanisms. 
The overall training procedure is shown in Algorithm~\ref{Alg:1}.

\begin{table}[!t]
\centering
\begin{tabularx}{\linewidth}{X}
\toprule
\textbf{Context:} \\
{$\text{Warplanes}_{instrument}$} \underline{pounded}
forward Iraqi positions in the hills overlooking 
Chamchamal, 35 kilometers ... \\
\midrule
\textbf{Generated Question:}
\\ 
What {$instrument$} was used in the attack in Iraqi positions? \\
\midrule
\textbf{Recovered Context:} \\
An $instrument$ was used to pound the Iraqi positions during the attack. \\
\bottomrule
\end{tabularx}
\caption{An example of inverse prompting mechanism from ACE. The recovered context is a basic rephrased description of the original context. More context-dependent questions will lead to higher similarity between recovered and original contexts.}
\label{tab:context reproduction}
\end{table}

\paragraph{Inverse Prompting Reward.}
The inverse prompting mechanism aims to recover the context from the question.
We assume that a better question is more context-dependent, and therefore, it is easier to recover the context to some extent.
Here, we developed an inverse prompting model to achieve context recovery, which is a text generation model that takes the following inputs:
\begin{equation}\label{eq:iprompt}
    p^{i}_{(t', q)} = \text{trigger}: []_{t'} \oplus \text{question}: []_{q}
\end{equation}
where $q$ is the question previously generated by Eq.~\ref{eq:qg}. 
The goal of inverse prompting is to recover the original context $c$; however, considering that it may contain information not appearing in the question, to ease the generation, we build a rephrased context $\hat{c}$ as a workaround.
Particularly, for each role in the ACE ontology, we chose one example question and manually constructed $\hat{c}$ (given an example shown in Table~\ref{tab:context reproduction}), and we used ChatGPT to obtain more question-rephrased context pairs by using a few-shot prompting. 
Then, we train the inverse prompting model to recover $\hat{c}$ from $p^{i}$. 
More details are shown in Appendix~\ref{inverseprompting-appendix}.

Finally, given a context $c$ and its recovered context $\hat{c}$, we utilize semantic similarity (SemSim) to evaluate the quality of recovery, denoted as $\operatorname{SemSim}(c, \hat{c})$, which provides an inverse prompting reward for further use.

\paragraph{Question Answering Reward.}
Intuitively, a good question should successfully guide a QA model to yield a correct answer.
Therefore, for each candidate question in $\mathcal{Q}$, we take it as the input of a QA model and generate a corresponding answer. For each question $q$ with standard answer $a$, we use the context overlap ratio (COR) to evaluate the predicted answer $\hat{a}$, which is obtained by:
\begin{equation}\label{eq:cor}
    \operatorname{COR}(a, \hat{a}) = \frac{|a \cap \hat{a}|}{\max\{|a|, |\hat{a}|\}},
\end{equation}
the $\operatorname{COR}$ score is calculated at the word level, and we use the tokenizer from NLTK\footnote{\href{https://www.nltk.org/}{https://www.nltk.org/}} for implementations. We then utilize $\operatorname{COR}(a, \hat{a})$ as the question-answering reward for further use. 

\paragraph{Refining with RL.}

\begin{algorithm}[t]
\caption{RL for QG Refinement}
\begin{algorithmic}[1]
\For{each $(c', t', r') \in $ training set $D$}
    \State Generate question set $\mathcal{Q}$ with $f_{SFT}$
    \For{each question $q \in \mathcal{Q}$}
        \State Generate recovered context $\hat{c}$
        \State Generate predicted answer $\hat{a}$
        \State Compute score using Eq.~\ref{eq:score} 
    \EndFor  
    \State \Return Reward score set $S_{\mathcal{Q}}$
    \If {Condition in Eq.~\ref{eq:selection} satisfied} 
    \State {\Return $(q_{+}, q_{-})$ as Eq.~\ref{eq:preferencepair}}
    \EndIf
\EndFor
\State Reward modeling $r(p, q)$ with Eq.~\ref{eq:rm}
\State PPO training with objective function Eq.~\ref{eq:ppo}
\State \Return RL-refined model $f_{RL}$
\end{algorithmic} 
\label{Alg:1}
\end{algorithm}

Typically, RL fine-tuning is employed subsequent to supervised fine-tuning for further refinement.
We introduce a reinforcement learning framework to refine the QG module.
Particularly, for each candidate question $q \in \mathcal{Q}$, we derive a combined score $S_q$ according to inverse prompting and question answering reward:
\begin{equation}\label{eq:score}
S_q = \lambda_{1}\operatorname{SemSim}(c, \hat{c}) + \lambda_{2}\operatorname{COR}(a, \hat{a}),
\end{equation}
the overall score of the question set $\mathcal{Q}$ can be represented as  $S_{\mathcal{Q}} = \{S_{q_{1}}, S_{q_{2}}, \ldots, S_{q_{n}}\}$.
Next, we select preferred/dispreferred question pair from the question set $\mathcal{Q}$ according to the following criteria:
\begin{equation}\label{eq:selection}
    \begin{cases}
    \max(S_{\mathcal{Q}}) > \alpha \\
    \max(S_{\mathcal{Q}}) - \min(S_{\mathcal{Q}}) > \beta, \\
    \end{cases}
\end{equation}
if the condition above is satisfied for $\mathcal{Q}$, we return the question pair with the highest and lowest score:
\begin{equation}\label{eq:preferencepair}
    (q_{+}, q_{-}) = (q_{\max(S_{\mathcal{Q}})}, q_{\min(S_{\mathcal{Q}})}),
\end{equation}
and then combine the question pair with their corresponding input $p$ to construct a preference dataset $D^{+/-}$. The reward modeling is minimizing the loss function for each $(p, q_{+}, q_{-})\in D^{+/-}$:
\begin{equation}\label{eq:rm}
\mathcal{L}_{\mathrm{RM}}=-\mathbb{E}\left[\log \left(\sigma\left(r\left(p,q^{+}\right)-r\left(p,q^{-}\right)\right)\right)\right],
\end{equation}
where $\sigma$ is the sigmoid function, $r$ is the score of question $q$ given $p$. The purpose of reward modeling is to get a reward function $r(p, q)$, where the higher the reward $r(p, q)$, the better the question $q$ is to the given input $p$. 
Denote the question generation model by supervised fine-tuned as $f_{SFT}$, the RL refining process is then to maximize the following objective function:
\begin{equation}\label{eq:ppo}
\mathcal{L}_{\mathrm{RL}}=\mathbb{E}\left[r\left(p, q\right)\right]-\mu \mathbb{E} \operatorname{KL}\left(f_{RL} \mid f_{SFT} \right),
\end{equation}
where $f_{RL}$ is the target model of the refinement, $\operatorname{KL}$ is Kullback-Leibler regularization, and $\mu> 0$ is the regularization parameter. This procedure enables a model to generate a better question with context-dependent and indicative refinement. Specifically, our RL framework utilizes the proximal policy optimization (PPO) algorithm~\cite{schulman-etal-2017-ppo}.

\subsection{Question Answering Module}
Finally, the event trigger, the target role, and the event context are given at the testing time. With an RL-refined model $f_{RL}$, we generate a question and use it as the prompt to trigger a QA model.
Considering that the best performance models are usually proprietary models, we do not train an additional QA model like in previous works~\cite{du-cardie-2020-event,lu-etal-2023-event} but directly use an off-the-shelf QA model such as LLaMA-2-13b-Chat or ChatGPT. 
We enumerate each role and get the answer as the EE results.
\section{Experimental Setups}\label{sec:exp setup}

\subsection{Datasets}

We conduct our experiments on the widely used dataset ACE 2005~\cite{doddington-etal-2004-automatic} and RAMS~\cite{ebner-etal-2020-multi}. ACE 2005 has 33 event types and 22 argument roles, which contains 599 documents crawled between 2003 and 2005 from various areas. We follow the same data split and preprocessing step as in the prior works~\cite{wadden-etal-2019-entity}. RAMS is a recently introduced dataset with document-level event argument extraction, which encompasses 9,124 annotated event mentions spanning 139 event types and 65 argument roles. More details of data preprocessing are given in Appendix~\ref{sec:appendix data-preprocessing}.

\subsection{Evaluation Settings}
In the ACE dataset, considering not all roles have related arguments in a given instance, we consider two settings for evaluations: 
(1) Practical evaluation: Only evaluate the questions whose target role has a corresponding argument (answerable questions). 
(2) Full evaluation: Evaluate the questions whose target role is all possible roles in ACE ontology (some of them are unanswerable questions)~\cite{rajpurkar-etal-2018-know}.

In the RAMS dataset, all the questions are in practical evaluation. For evaluation metrics, considering that exactly matching the QA model's response with the standard answer is difficult for an off-the-shelf model, we selected several metrics of varying degrees. 

The metric considers: 1) Exact Match Accuracy (EM), a predicted answer is considered correct only if all of it exactly matches the standard answer. 
2) Context Overlap Ratio (COR), which can be computed with Eq.~\ref{eq:cor}. 
3) Semantic Similarity (SemSim),  which can evaluate the detailed response.

\subsection{Implementations}
In our implementations of the QG module, we use LLaMA-2-7b~\cite{touvron-etal-2023-llama2} as the base model; the learning rate is set as 5e-5, selected from the interval [1e-5, 1e-4].
For the QA model, we adopt the freeze (off-the-shelf) model in both training and testing, LLaMA-2-13b-Chat, and we also adopt a 5-shot in prompting to help the QA model's understanding. All the question-answering processes use the same few-shot examples and this setting.
The selection of preferred/dispreferred pairs is selected according to condition Eq.~\ref{eq:selection} and reward score Eq.~\ref{eq:score}. In practice, to balance the model's performance, we set $\lambda_{1}$ to 0.3 and $\lambda_{2}$ to 0.7 in Eq.~\ref{eq:score}. In Eq.~\ref{eq:selection}, the $\alpha$ is set to 0.65, and $\beta$ is set to 0.5.
The discussion of model selection and the details of training and hyperparameters are given in Appendix~\ref{sec:appendix implementation}.

\subsection{Baselines}
We divide baselines into three groups:
1) Template-based methods: RCEE~\cite{liu-etal-2020-event}, which uses a simple template such as ``WH is the [role]?'', denoted as Simple-Q\footnote{Q stands for Question.}. EEQA~\cite{du-cardie-2020-event}, which introduces two types of questions: a template that incorporates the trigger and Simple-Q, denoted as Standard-Q. Moreover, the question generated by the descriptions of each argument role is provided in the ACE annotation guidelines for events, denoted as Guideline-Q. Back-Translation, which back-translate the Standard-Q denoted as Back-Trans-Q. QGA-EE~\cite{lu-etal-2023-event}, which designed a dynamic template for different event type and their corresponding role, denoted as Dynamic-Q. 
2) Supervised fine-tuning-based methods: Which trains a QG model via supervised fine-tuning using the above templates\footnote{Using different templates in Section~\ref{sec:qg}.}. They are denoted as SFT (\textit{Template}) respectively.
3) In-context learning-based methods: These perform 0-shot and 5-shot on LLaMA-2-13b-Chat and GPT-4, respectively. The few-shot example will be shown in Appendix~\ref{sec:appendix fewshots}. 
Noting that the above methods only studied and designed the template questions on ACE, considering the generalizability of the method above, we adopt Standard-Q and Back-Trans-Q as the RAMS experimental baselines. Our proposed method, \textbf{R}einforcement \textbf{L}earning-based \textbf{Q}uestion \textbf{G}eneration refinement, is called \textbf{RLQG}. Specifically, we train our model on the ACE dataset based on the most contextualized template, Dynamic-Q. On the RAMS dataset, we train the model based on a more natural question, Back-Trans-Q. We will discuss the influence of different template starting points in Section~\ref{sec:qg-ablation}.
\begin{table*}[!t]
  \centering
  \setlength\tabcolsep{7pt}
  \begin{tabular}{lcccccc}
    \toprule
    \multicolumn{1}{c}{\multirow{2}{*}{\textbf{Methods}}} & \multicolumn{3}{c}{\textbf{Practical Eval.}} & \multicolumn{3}{c}{\textbf{Full Eval.}}  \\
    \cmidrule(lr){2-4} \cmidrule(lr){5-7}
    & {EM} & {COR} & {SemSim} & {EM} & {COR} & {SemSim}\\
    \midrule
    \multicolumn{2}{l}{\textit{Template}} \\
    Simple-Q~\cite{liu-etal-2020-event} & 35.41 & 40.23 & 60.93 & 14.38 & 16.17 & 24.55 \\
    Standard-Q~\cite{du-cardie-2020-event} & 37.42 & 43.87 & 63.70 & 15.60 & 17.36 & 25.92\\
    Back-Trans-Q & 36.13 & 41.39 & 62.41 & 15.14 & 16.67 & 25.28\\
    Guideline-Q~\cite{du-cardie-2020-event} & 38.51 & 45.28 & 65.54 & 17.61 & 19.96 & 28.14 \\
    Dynamic-Q~\cite{lu-etal-2023-event}  & 38.70 & 45.79 & 65.55 & 20.45 & 23.12 & 30.79 \\
    \midrule
    \multicolumn{2}{l}{\textit{Supervised Fine-tuning}} \\
    \text{SFT} (\textit{Standard}) & 37.63 & 42.95 & 62.36 & 15.31 & 17.13 & 25.72  \\
    \text{SFT} (\textit{Back-Trans}) & 38.24 & 43.56 & 64.11 & 17.47 & 18.90 & 27.32  \\
    \text{SFT} (\textit{Guideline}) & 38.62 & 44.69 & 64.66 & 17.33 & 19.61 & 27.77  \\
    \text{SFT} (\textit{Dynamic}) & 39.31 & 46.78 & 66.24 & 20.35 & 23.05 & 30.53  \\
    \midrule
    \multicolumn{2}{l}{\textit{In-context learning}} \\
    LLaMA-2-13b-Chat (\textit{0shot}) & 1.21 & 3.50 & 35.88 & 0.43 & 1.25 & 21.78 \\ 
    LLaMA-2-13b-Chat (\textit{5shot}) & 27.97 & 33.04 & 53.69 & 13.01 & 14.93 & 23.54 \\
    GPT-4 (\textit{0shot}) & 28.97 & 35.83 & 57.90 & 11.14 & 13.54 & 23.35 \\
    GPT-4 (\textit{5shot}) & 39.24 & 47.59 & 65.92 & 16.32 & 19.37 & 27.46 \\
    \midrule
    \textbf{RLQG (Ours)} & \textbf{41.39} & \textbf{48.58} & \textbf{67.94} & \textbf{21.71} & \textbf{24.19} & \textbf{31.80} \\
    \bottomrule
\end{tabular}
\caption{Event extraction results with Practical Evaluation and Full Evaluation on the ACE test dataset, where EM, COR, and SemSim indicate exact match accuracy, context overlap ratio, and semantic similarity, respectively.}
\label{tab:ace}
\end{table*}
\section{Experimental Results}
In this section, we present the experimental results divided by results with full training resources on ACE and RAMS and results on the data-scarce scenarios in ACE.

\subsection{Results with Full Training Resource}

Table~\ref{tab:ace} gives the results on ACE with full training resources. 
From the results: 1) Above all baseline methods we compared, it is evident that our method surpasses all the baselines in terms of three given metrics and evaluation settings. 
Specifically, our method outperforms the second-best method SFT (\textit{Dynamic}) by 2.08\% and outperforms the template starting point Dynamic-Q by 2.69\% in EM, with practical evaluation. 
We attribute the results to the RL refinement with rewards of inverse prompting and question answering in our framework, which helps our question become more context-dependent and indicative than the other method. 
We also surpassed the GPT-4 (\textit{5shot}) in evaluation, demonstrating the powerful potential of our model compared to the proprietary model. 
2) By comparing different baselines, we explore that the quality of template questions determines the ability of the SFT model for question generation. Also, a higher quality template as the starting point to train the SFT model will improve the SFT process. The results with in-context learning methods show that the model's ability affects the performance, and the few-shot example also brings a guideline on better question generation.

\begin{table}[t]
  \centering  
  \begin{tabular}{lcccc}
    \toprule
    \textbf{Methods} & EM & COR & SemSim\\
    \midrule
    \multicolumn{2}{l}{\textit{Template}} \\
      Standard-Q & 17.65 & 23.02 & 47.96 \\
      Back-Trans-Q & 16.45 & 21.47 & 46.43 \\
    \midrule
    \multicolumn{2}{l}{\textit{Supervised Fine-tuning}} \\
      \text{SFT} (\textit{Standard}) & 18.10 & 23.84 & 48.79 \\
      \text{SFT} (\textit{Back-Trans}) & 18.29 & 24.11 & 49.32 \\
    \midrule
      \textbf{RLQG (Ours)} & \textbf{19.61} & \textbf{25.43} & \textbf{50.69} \\
    \bottomrule
  \end{tabular}
\caption{Event extraction results on the RAMS test dataset with practical evaluation.}
\label{tab:rams}
\end{table}

Table~\ref{tab:rams} gives the evaluation results of the RAMS test set.
From the results: 1) It is also evident that our method RLQG outperforms the comparative baseline on all metrics, which obtains 1.32\% improvement on EM compared to the second-best model SFT (\textit{Back-Trans}), and also outperforms the best template method Standard-Q by 1.96\%.
2) Compared to the static form Standard-Q, the improvement of fine-tuning to the Back-Trans-Q is more significant; we conclude that the model improves more on natural questions.

\subsection{Results in Data-Scarce Scenarios}

\begin{figure}[!t]
  \centering
  \begin{subfigure}{.24\textwidth}
    \centering
    \includegraphics[width=.9\linewidth]{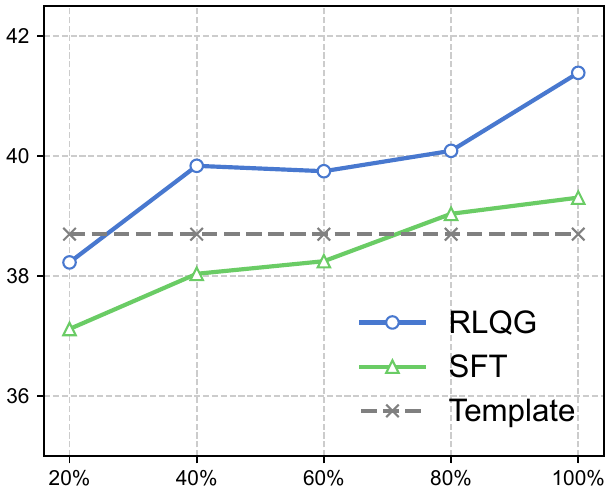}
    \caption{Exact Match Accuracy}
    \label{fig:acc}
  \end{subfigure}%
  \begin{subfigure}{.24\textwidth}
    \centering
    \includegraphics[width=.9\linewidth]{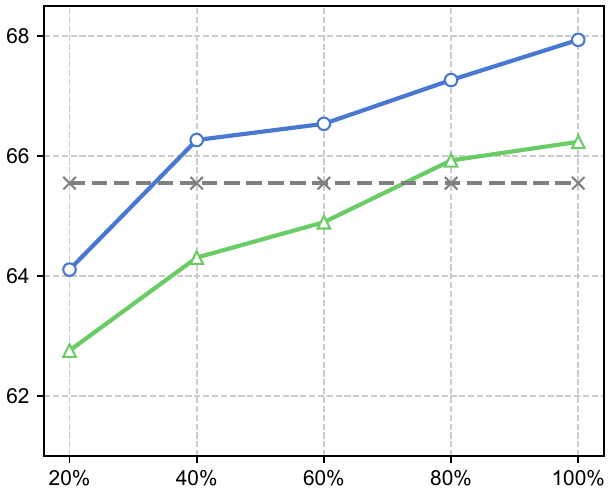} 
    \caption{Semantic Similarity}
    \label{fig:simil}
  \end{subfigure}
  \caption{Experimental results in the ACE dataset for the data-scarce scenario. The x-axis represents different ratios of training data, y-axis is the value of the metric.}
\end{figure}

To explore the performance of our model in the data-scarcity settings, we randomly choose x\% number of dynamic templates~\cite{lu-etal-2023-event} to fill in and use these questions as our training data. 
The comparative baseline is the template-based method Dynamic-Q and supervised fine-tuning-based method SFT (\textit{Dynamic}). 
The results shown in Figure~\ref{fig:acc} and~\ref{fig:simil} represent their performance on the different metrics in the data-scarce scenarios.

From the results, it is significant that our method outperforms the SFT method in data scarcity, especially when the data is limited to around 40\% to 60\%; our method opens up a big gap with the fine-tuning method. In addition, we can outperform the full-annotated template by using only 40\% of training data, performing a good adaptation with a data-scarce scenario and budget limitations. As further explanation, the SFT method can only learn from annotated data, and when training data is limited, it struggles with generalization issues, resulting in poor performance. In contrast, our method is more data-efficient because it combines two mechanisms: 1) The QG model, combined with the beam search mechanism, can generate more diverse questions, potentially providing more supervision signals. 2) More importantly, our RL refinement mechanism assists in ranking such questions and identifying better ones for optimization, resulting in improved generalization ability. By combining the two mechanisms, our method can make the best use of limited data while producing superior results.
\section{Further Discussion}
If not explicitly stated, further discussion of our method's details focuses on the representative baseline with the template-based method Dynamic-Q and its corresponding SFT model. The experiments are conducted on the ACE dataset with practical evaluation.

\subsection{Ablations on QG Architecture}\label{sec:qg-ablation}
\paragraph{Template Starting Point.}
We turn our method's starting point on the template question that with the most minor human intervention: Standard-Q, which can obtained by simply concatenating the interrogative pronoun, the role, and the trigger word. The experiments are conducted in both datasets on the metric EM. 
Table~\ref{tab:template-ablation} shows the corresponding results, in which our method outperforms the template-based and supervised fine-tuning-based method in both the ACE and RAMS datasets. This is a competitive result that our method can also achieve good performance even if there is only rarely manual intervention.

\begin{table}[t]
\setlength\tabcolsep{12pt}
\centering
\begin{tabular}{lcc}
\hline
\textbf{Methods} & \textbf{ACE} & \textbf{RAMS} \\ 
\hline
Standard-Q & 37.42 & 17.65   \\
\text{SFT} (\textit{Standard}) & 37.63 & 18.10  \\
\textbf{RLQG} (\textit{Standard}) & \textbf{39.29} & \textbf{19.23}  \\
\hline
\end{tabular}
\caption{Ablation study of the different template as the starting point on exact match accuracy}
\label{tab:template-ablation}
\end{table} 

\paragraph{Rewards of Two Mechanism.}
We conduct the ablation study on inverse prompting (IP) and question answering (QA) rewards; Table~\ref{tab:lambda-ablation} lists different variations of our proposed method.
When removing the IP reward, the exact match accuracy (EM) decreased by 1.18\%, and the COR and SemSim decreased by 1.09\% and 0.98\%, respectively. And without QA reward, the decrease became 1.53\%, 1.41\%, and 1.06\% to these three metric. These results indicate that the indicative is prior in question generation. Overall, each component of the RLQG method plays a crucial role in achieving good performance, as their removal resulted in decreased performance across all metrics.

\begin{table}[t]
\setlength\tabcolsep{5pt}
\centering
\begin{tabular}{lccc}
\hline
\textbf{Method} & \textbf{EM} & \textbf{COR} & \textbf{SemSim} \\ 
\hline
\textbf{RLQG} & \textbf{41.39} & \textbf{48.58} & \textbf{67.94}  \\
-w/o \textit{IP reward} & 40.21 & 47.49 & 66.96 \\
-w/o \textit{QA reward} & 39.86 & 47.17 & 66.88 \\
\hline
\end{tabular}
\caption{Results of ablation studies for removing different reward.}
\label{tab:lambda-ablation}
\end{table}

\subsection{Ablation on QA architectures}
\begin{figure}[t]
  \centering
  \begin{subfigure}{.24\textwidth}
    \centering
    \includegraphics[width=.93\linewidth]{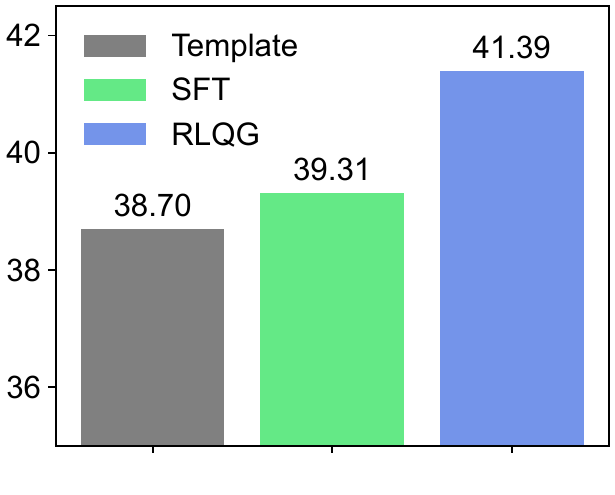}
    \caption{}
  \end{subfigure}%
  \begin{subfigure}{.24\textwidth}
    \centering
    \includegraphics[width=.93\linewidth]{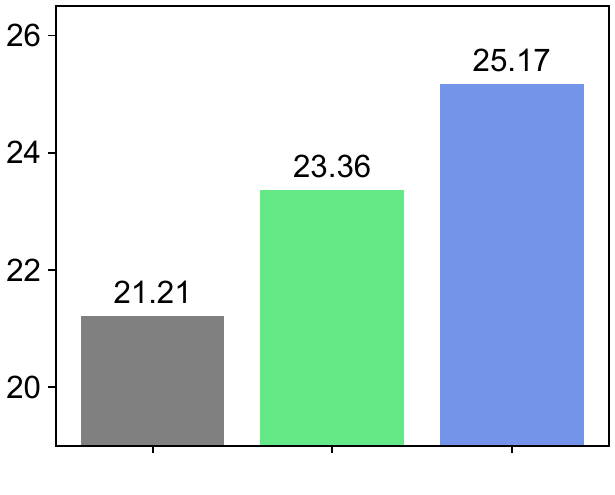} 
    \caption{}
  \end{subfigure}
  \begin{subfigure}{.24\textwidth}
    \centering
    \includegraphics[width=.93\linewidth]{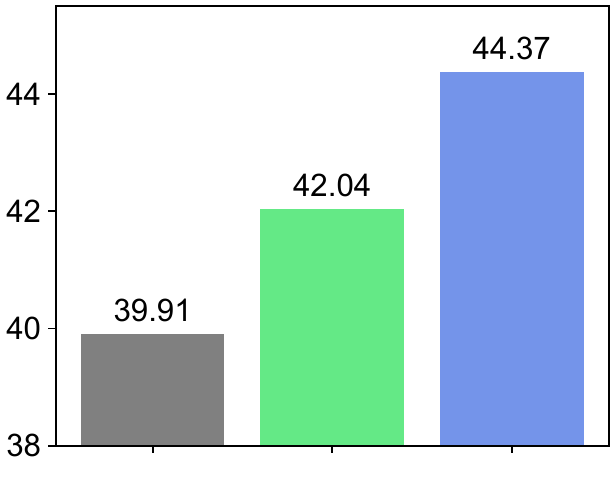}
    \caption{}
    \label{fig:gpt-intro1}
  \end{subfigure}%
  \begin{subfigure}{.24\textwidth}
    \centering
    \includegraphics[width=.93\linewidth]{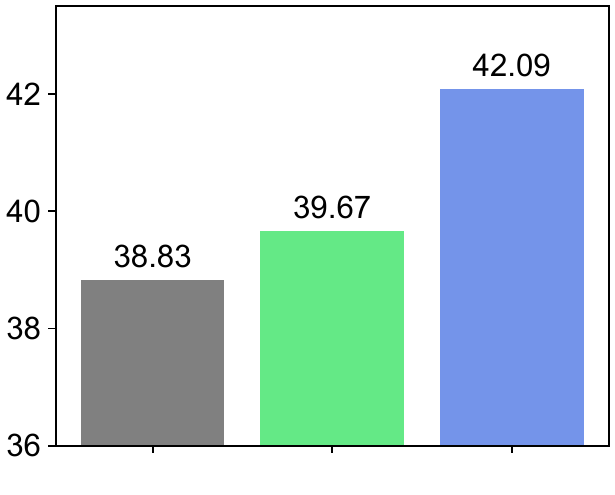} 
    \caption{}
    \label{fig:gpt-intro2}
  \end{subfigure}
  \caption{The performance with exact match accuracy on different QA model (a) LLaMA-2-13b-Chat (\textit{5shot}) (identical to previous settings) (b) LLaMA-2-13b-Chat (\textit{0shot}) (c) GPT-4 (\textit{5shot}) (d) GPT-4 (\textit{0shot})}
  \label{fig:QA-ablation}
\end{figure}

This section discusses the impact of different QA models in our method, we explore the universal evaluation of the question quality by adopting different QA modules. 
As a prompting aspect, we compare the QA module with 5-shot and 0-shot. Moreover, considering the model aspect, we utilize GPT-4 as a comparative QA model.
As shown in Figure~\ref{fig:QA-ablation}, our method outperforms the other two methods with different QA models in all cases. The basic capacity of the QA model will lead to the diversity of the performance on EAE. However, our method represents suitability for different selections of QA models.

\subsection{Case Study}
As expected, our method will be a better question regarding fluency, context dependence, and generalizability. We present an intuitive analysis of a case study. As shown in Table~\ref{tab:case study}, we select an example in the ACE05 test dataset. The target of these questions is to trigger the QA model to extract the answer ``McCarthy''. Obviously, the question generated by RLQG get the correct answer and is most similar to human-written questions.

\begin{table}[t]
\centering
\begin{tabularx}{\linewidth}{X}
\toprule
\textbf{Context:} \\
Former senior banker $\text{Callum McCarthy}_{person}$ \textbf{begins} what is one of the most important jobs in London's financial world in September, when incumbent Howard Davies steps down. \\
\midrule
\textbf{Template Question:} \\
Who is the employee?\\
\textbf{Answer:} Davies \textcolor{red}{\ding{55}}\\
\midrule
\textbf{SFT Question:} \\
Who was hired by banker? \\
\textbf{Answer:} Howard Davies \textcolor{red}{\ding{55}} \\
\midrule
\textbf{RLQG Question:} \\
Who was hired as one of the most important jobs? \\
\textbf{Answer:} Callum McCarthy \textcolor{green}{\ding{51}} \\

\midrule
\textbf{Human-Written Question:} \\
Who was the former senior banker that began an important job in September?\\
\textbf{Answer:} Callum McCarthy \textcolor{green}{\ding{51}}\\

\bottomrule
\end{tabularx}
\caption{An instance of the ACE test set, we use different methods to generate a corresponding question on semantic role \textit{person}; our method gets the correct answer and gets close to a human-written question.}
\label{tab:case study}
\end{table}


\section{Conclusion}
Event extraction (EE) has evolved from traditional classification into question-answering-based methods (QA-based EE). These methods emphasize the design of quality questions to guide QA models for better answers. In this paper, we introduce a reinforcement learning framework that aims to produce context-dependent, fluently phrased questions that are generalizable and indicative enough to QA models, addressing the challenges posed by rigid, template-based questions. Our methodology demonstrates improved performance on ACE and RAMS benchmarks, particularly in data-scarce scenarios. It highlights the method's efficacy in generating effective questions for QA-based EE without extensive manual intervention.

\section{Limitations}
Two primary limitations are acknowledged in this study. Firstly, most existing QA-based EE approaches assume known triggers, effectively overlooking the impact of trigger identification. This study follows this assumption but plans to incorporate event detection (ED) in future work for a comprehensive approach to the event extraction (EE) task. Secondly, the method's generalizability to real-world scenarios remains to be determined, as it has only been evaluated on standard datasets. The complexity, diversity, and potential noise of real-world data call for further validation to confirm the method's effectiveness in practical applications. Additionally, our research raises no ethical issues because it focuses solely on the technical aspects of a normal information extraction problem.


\nocite{*}
\bibliography{custom}

\begin{thebibliography}{48}
\expandafter\ifx\csname natexlab\endcsname\relax\def\natexlab#1{#1}\fi

\bibitem[{Ahn(2006)}]{ahn-2006-stages}
David Ahn. 2006.
\newblock \href {https://aclanthology.org/W06-0901} {The stages of event extraction}.
\newblock In \emph{ACL 2006 Workshop on Annotating and Reasoning about Time and Events (ACL)}.

\bibitem[{Bai et~al.(2023)Bai, Bai, Chu, Cui, Dang, Deng, Fan, Ge, Han, Huang et~al.}]{bai2023qwen}
Jinze Bai, Shuai Bai, Yunfei Chu, Zeyu Cui, Kai Dang, Xiaodong Deng, Yang Fan, Wenbin Ge, Yu~Han, Fei Huang, et~al. 2023.
\newblock \href {https://arxiv.org/pdf/2309.16609} {Qwen technical report}.
\newblock \emph{arXiv preprint arXiv:2309.16609}.

\bibitem[{Chen et~al.(2023)Chen, Ma, Wang, and Cohen}]{chen-etal-2023-program}
Wenhu Chen, Xueguang Ma, Xinyi Wang, and William~W. Cohen. 2023.
\newblock \href {https://openreview.net/forum?id=YfZ4ZPt8zd} {Program of thoughts prompting: Disentangling computation from reasoning for numerical reasoning tasks}.
\newblock \emph{Transactions on Machine Learning Research (TMLR)}.

\bibitem[{Chen et~al.(2015)Chen, Xu, Liu, Zeng, and Zhao}]{chen-etal-2015-event}
Yubo Chen, Liheng Xu, Kang Liu, Daojian Zeng, and Jun Zhao. 2015.
\newblock \href {https://aclanthology.org/P15-1017} {Event extraction via dynamic multi-pooling convolutional neural networks}.
\newblock In \emph{Association for Computational Linguistics and International Joint Conference on Natural Language Processing (ACL-IJCNLP)}.

\bibitem[{Chen et~al.(2024)Chen, Deng, Yuan, Ji, and Gu}]{chen-etal-2024-selfplay}
Zixiang Chen, Yihe Deng, Huizhuo Yuan, Kaixuan Ji, and Quanquan Gu. 2024.
\newblock \href {https://arxiv.org/pdf/2401.01335} {Self-play fine-tuning converts weak language models to strong language models}.
\newblock \emph{arXiv preprint arXiv:2401.01335}.

\bibitem[{Christiano et~al.(2017)Christiano, Leike, Brown, Martic, Legg, and Amodei}]{christiano-etal-2017-rlhf}
Paul~F Christiano, Jan Leike, Tom Brown, Miljan Martic, Shane Legg, and Dario Amodei. 2017.
\newblock \href {https://proceedings.neurips.cc/paper_files/paper/2017/file/d5e2c0adad503c91f91df240d0cd4e49-Paper.pdf} {Deep reinforcement learning from human preferences}.
\newblock In \emph{Advances in Neural Information Processing Systems (NeurIPS)}.

\bibitem[{Doddington et~al.(2004)Doddington, Mitchell, Przybocki, Ramshaw, Strassel, and Weischedel}]{doddington-etal-2004-automatic}
George Doddington, Alexis Mitchell, Mark Przybocki, Lance Ramshaw, Stephanie Strassel, and Ralph Weischedel. 2004.
\newblock \href {http://www.lrec-conf.org/proceedings/lrec2004/pdf/5.pdf} {The automatic content extraction ({ACE}) program {--} tasks, data, and evaluation}.
\newblock In \emph{International Conference on Language Resources and Evaluation (LREC)}.

\bibitem[{Du and Cardie(2020)}]{du-cardie-2020-event}
Xinya Du and Claire Cardie. 2020.
\newblock \href {https://aclanthology.org/2020.emnlp-main.49} {Event extraction by answering (almost) natural questions}.
\newblock In \emph{Empirical Methods in Natural Language Processing (EMNLP)}.

\bibitem[{Ebner et~al.(2020)Ebner, Xia, Culkin, Rawlins, and Van~Durme}]{ebner-etal-2020-multi}
Seth Ebner, Patrick Xia, Ryan Culkin, Kyle Rawlins, and Benjamin Van~Durme. 2020.
\newblock \href {https://aclanthology.org/2020.acl-main.718} {Multi-sentence argument linking}.
\newblock In \emph{Association for Computational Linguistics (ACL)}.

\bibitem[{Gao et~al.(2021)Gao, Fisch, and Chen}]{gao-etal-2021-making}
Tianyu Gao, Adam Fisch, and Danqi Chen. 2021.
\newblock \href {https://aclanthology.org/2021.acl-long.295} {Making pre-trained language models better few-shot learners}.
\newblock In \emph{Association for Computational Linguistics and International Joint Conference on Natural Language Processing (ACL-IJCNLP)}.

\bibitem[{Hong et~al.(2024)Hong, Yuan, Chen, Zhang, Huang, and Huang}]{hong2024knowledgetosql}
Zijin Hong, Zheng Yuan, Hao Chen, Qinggang Zhang, Feiran Huang, and Xiao Huang. 2024.
\newblock \href {https://arxiv.org/pdf/2402.11517} {Knowledge-to-sql: Enhancing sql generation with data expert llm}.
\newblock \emph{arXiv preprint arXiv:2402.11517}.

\bibitem[{Hu et~al.(2021)Hu, Shen, Wallis, Allen-Zhu, Li, Wang, Wang, and Chen}]{hu-etal-2021-lora}
Edward~J Hu, Yelong Shen, Phillip Wallis, Zeyuan Allen-Zhu, Yuanzhi Li, Shean Wang, Lu~Wang, and Weizhu Chen. 2021.
\newblock \href {https://arxiv.org/pdf/2106.09685} {Lora: Low-rank adaptation of large language models}.
\newblock \emph{arXiv preprint arXiv:2106.09685}.

\bibitem[{Kojima et~al.()Kojima, Gu, Reid, Matsuo, and Iwasawa}]{kojima-etal-2022-zeroshots}
Takeshi Kojima, Shixiang~(Shane) Gu, Machel Reid, Yutaka Matsuo, and Yusuke Iwasawa.
\newblock \href {https://proceedings.neurips.cc/paper_files/paper/2022/file/8bb0d291acd4acf06ef112099c16f326-Paper-Conference.pdf} {Large language models are zero-shot reasoners}.
\newblock In \emph{Advances in Neural Information Processing Systems (NeurIPS)}.

\bibitem[{Lester et~al.(2021)Lester, Al-Rfou, and Constant}]{lester-etal-2021-power}
Brian Lester, Rami Al-Rfou, and Noah Constant. 2021.
\newblock \href {https://aclanthology.org/2021.emnlp-main.243} {The power of scale for parameter-efficient prompt tuning}.
\newblock In \emph{Empirical Methods in Natural Language Processing (EMNLP)}.

\bibitem[{Lewis et~al.(2020)Lewis, Perez, Piktus, Petroni, Karpukhin, Goyal, K\"{u}ttler, Lewis, Yih, Rockt\"{a}schel, Riedel, and Kiela}]{lewis-etal-2020-rag}
Patrick Lewis, Ethan Perez, Aleksandra Piktus, Fabio Petroni, Vladimir Karpukhin, Naman Goyal, Heinrich K\"{u}ttler, Mike Lewis, Wen-tau Yih, Tim Rockt\"{a}schel, Sebastian Riedel, and Douwe Kiela. 2020.
\newblock \href {https://proceedings.neurips.cc/paper_files/paper/2020/file/6b493230205f780e1bc26945df7481e5-Paper.pdf} {Retrieval-augmented generation for knowledge-intensive nlp tasks}.
\newblock In \emph{Advances in Neural Information Processing Systems (NeurIPS)}.

\bibitem[{Li et~al.(2020)Li, Peng, Chen, Wang, Pan, Lyu, and Zhu}]{li-etal-2020-event}
Fayuan Li, Weihua Peng, Yuguang Chen, Quan Wang, Lu~Pan, Yajuan Lyu, and Yong Zhu. 2020.
\newblock \href {https://aclanthology.org/2020.findings-emnlp.73} {Event extraction as multi-turn question answering}.
\newblock In \emph{Findings of Empirical Methods in Natural Language Processing (EMNLP)}.

\bibitem[{Li et~al.(2013)Li, Ji, and Huang}]{li-etal-2013-joint}
Qi~Li, Heng Ji, and Liang Huang. 2013.
\newblock \href {https://aclanthology.org/P13-1008} {Joint event extraction via structured prediction with global features}.
\newblock In \emph{Association for Computational Linguistics (ACL)}.

\bibitem[{Li et~al.(2021)Li, Ji, and Han}]{li-etal-2021-document}
Sha Li, Heng Ji, and Jiawei Han. 2021.
\newblock \href {https://aclanthology.org/2021.naacl-main.69} {Document-level event argument extraction by conditional generation}.
\newblock In \emph{North American Chapter of the Association for Computational Linguistics: Human Language Technologies (NAACL-HLT)}.

\bibitem[{Li and Liang(2021)}]{li-liang-2021-prefix}
Xiang~Lisa Li and Percy Liang. 2021.
\newblock \href {https://aclanthology.org/2021.acl-long.353} {Prefix-tuning: Optimizing continuous prompts for generation}.
\newblock In \emph{Association for Computational Linguistics and International Joint Conference on Natural Language Processing (ACL-IJCNLP)}.

\bibitem[{Liu et~al.(2020)Liu, Chen, Liu, Bi, and Liu}]{liu-etal-2020-event}
Jian Liu, Yubo Chen, Kang Liu, Wei Bi, and Xiaojiang Liu. 2020.
\newblock \href {https://aclanthology.org/2020.emnlp-main.128} {Event extraction as machine reading comprehension}.
\newblock In \emph{Empirical Methods in Natural Language Processing (EMNLP)}.

\bibitem[{Liu et~al.(2021)Liu, Chen, and Xu}]{liu-etal-2021-machine}
Jian Liu, Yufeng Chen, and Jinan Xu. 2021.
\newblock \href {https://aclanthology.org/2021.emnlp-main.214} {Machine reading comprehension as data augmentation: A case study on implicit event argument extraction}.
\newblock In \emph{Empirical Methods in Natural Language Processing (EMNLP)}.

\bibitem[{Liu et~al.(2023)Liu, Yuan, Fu, Jiang, Hayashi, and Neubig}]{liu-etal-2023-prompt}
Pengfei Liu, Weizhe Yuan, Jinlan Fu, Zhengbao Jiang, Hiroaki Hayashi, and Graham Neubig. 2023.
\newblock \href {https://doi.org/10.1145/3560815} {Pre-train, prompt, and predict: A systematic survey of prompting methods in natural language processing}.
\newblock \emph{ACM Computing Surveys}.

\bibitem[{Lu et~al.(2023)Lu, Ran, Tetreault, and Jaimes}]{lu-etal-2023-event}
Di~Lu, Shihao Ran, Joel Tetreault, and Alejandro Jaimes. 2023.
\newblock \href {https://aclanthology.org/2023.acl-short.143} {Event extraction as question generation and answering}.
\newblock In \emph{Association for Computational Linguistics (ACL)}.

\bibitem[{Lyu et~al.(2021)Lyu, Zhang, Sulem, and Roth}]{lyu-etal-2021-zero}
Qing Lyu, Hongming Zhang, Elior Sulem, and Dan Roth. 2021.
\newblock \href {https://aclanthology.org/2021.acl-short.42} {Zero-shot event extraction via transfer learning: {C}hallenges and insights}.
\newblock In \emph{Association for Computational Linguistics and International Joint Conference on Natural Language Processing (ACL-IJCNLP)}.

\bibitem[{Ma et~al.(2022)Ma, Wang, Cao, Li, Chen, Wang, and Shao}]{ma-etal-2022-prompt}
Yubo Ma, Zehao Wang, Yixin Cao, Mukai Li, Meiqi Chen, Kun Wang, and Jing Shao. 2022.
\newblock \href {https://aclanthology.org/2022.acl-long.466} {{P}rompt for extraction? {PAIE}: {P}rompting argument interaction for event argument extraction}.
\newblock In \emph{Association for Computational Linguistics (ACL)}.

\bibitem[{Nguyen et~al.(2016)Nguyen, Cho, and Grishman}]{nguyen-etal-2016-joint-event}
Thien~Huu Nguyen, Kyunghyun Cho, and Ralph Grishman. 2016.
\newblock \href {https://aclanthology.org/N16-1034} {Joint event extraction via recurrent neural networks}.
\newblock In \emph{North American Chapter of the Association for Computational Linguistics: Human Language Technologies (NAACL-HLT)}.

\bibitem[{OpenAI(2023)}]{openai-2023-gpt}
OpenAI. 2023.
\newblock \href {https://arxiv.org/pdf/2303.08774} {Gpt-4 technical report}.
\newblock \emph{arXiv preprint arXiv:2303.08774}.

\bibitem[{Ouyang et~al.(2022)Ouyang, Wu, Jiang, Almeida, Wainwright, Mishkin, Zhang, Agarwal, Slama, Ray, Schulman, Hilton, Kelton, Miller, Simens, Askell, Welinder, Christiano, Leike, and Lowe}]{long-etal-2022-rlhff}
Long Ouyang, Jeffrey Wu, Xu~Jiang, Diogo Almeida, Carroll Wainwright, Pamela Mishkin, Chong Zhang, Sandhini Agarwal, Katarina Slama, Alex Ray, John Schulman, Jacob Hilton, Fraser Kelton, Luke Miller, Maddie Simens, Amanda Askell, Peter Welinder, Paul~F Christiano, Jan Leike, and Ryan Lowe. 2022.
\newblock \href {https://proceedings.neurips.cc/paper_files/paper/2022/file/b1efde53be364a73914f58805a001731-Paper-Conference.pdf} {Training language models to follow instructions with human feedback}.
\newblock In \emph{Advances in Neural Information Processing Systems (NeurIPS)}.

\bibitem[{Radford et~al.(2019)Radford, Wu, Child, Luan, Amodei, Sutskever et~al.}]{radford-etal-2019-unsupervisedmultitask}
Alec Radford, Jeffrey Wu, Rewon Child, David Luan, Dario Amodei, Ilya Sutskever, et~al. 2019.
\newblock \href {https://d4mucfpksywv.cloudfront.net/better-language-models/language_models_are_unsupervised_multitask_learners.pdf} {Language models are unsupervised multitask learners}.
\newblock \emph{OpenAI blog}.

\bibitem[{Rafailov et~al.(2023)Rafailov, Sharma, Mitchell, Manning, Ermon, and Finn}]{rafailov-etal-2023-direct}
Rafael Rafailov, Archit Sharma, Eric Mitchell, Christopher~D Manning, Stefano Ermon, and Chelsea Finn. 2023.
\newblock \href {https://openreview.net/forum?id=HPuSIXJaa9} {Direct preference optimization: Your language model is secretly a reward model}.
\newblock In \emph{Advances in Neural Information Processing Systems (NeurIPS)}.

\bibitem[{Raffel et~al.(2020)Raffel, Shazeer, Roberts, Lee, Narang, Matena, Zhou, Li, and Liu}]{raffel-etal-2020-t5}
Colin Raffel, Noam Shazeer, Adam Roberts, Katherine Lee, Sharan Narang, Michael Matena, Yanqi Zhou, Wei Li, and Peter~J. Liu. 2020.
\newblock \href {http://jmlr.org/papers/v21/20-074.html} {Exploring the limits of transfer learning with a unified text-to-text transformer}.
\newblock \emph{The Journal of Machine Learning Research (JMLR)}.

\bibitem[{Rajpurkar et~al.(2018)Rajpurkar, Jia, and Liang}]{rajpurkar-etal-2018-know}
Pranav Rajpurkar, Robin Jia, and Percy Liang. 2018.
\newblock \href {https://aclanthology.org/P18-2124} {Know what you don{'}t know: Unanswerable questions for {SQ}u{AD}}.
\newblock In \emph{Association for Computational Linguistics (ACL)}.

\bibitem[{Rajpurkar et~al.(2016)Rajpurkar, Zhang, Lopyrev, and Liang}]{rajpurkar-etal-2016-squad}
Pranav Rajpurkar, Jian Zhang, Konstantin Lopyrev, and Percy Liang. 2016.
\newblock \href {https://aclanthology.org/D16-1264} {{SQ}u{AD}: 100,000+ questions for machine comprehension of text}.
\newblock In \emph{Empirical Methods in Natural Language Processing (EMNLP)}.

\bibitem[{Schulman et~al.(2017)Schulman, Wolski, Dhariwal, Radford, and Klimov}]{schulman-etal-2017-ppo}
John Schulman, Filip Wolski, Prafulla Dhariwal, Alec Radford, and Oleg Klimov. 2017.
\newblock \href {https://arxiv.org/pdf/1707.06347} {Proximal policy optimization algorithms}.
\newblock \emph{arXiv preprint arXiv:1707.06347}.

\bibitem[{Shin et~al.(2020)Shin, Razeghi, Logan~IV, Wallace, and Singh}]{shin-etal-2020-autoprompt}
Taylor Shin, Yasaman Razeghi, Robert~L. Logan~IV, Eric Wallace, and Sameer Singh. 2020.
\newblock \href {https://aclanthology.org/2020.emnlp-main.346} {{A}uto{P}rompt: {E}liciting {K}nowledge from {L}anguage {M}odels with {A}utomatically {G}enerated {P}rompts}.
\newblock In \emph{Empirical Methods in Natural Language Processing (EMNLP)}.

\bibitem[{Tang et~al.(2022)Tang, Li, Zhao, and Wen}]{tang-etal-2022-context}
Tianyi Tang, Junyi Li, Wayne~Xin Zhao, and Ji-Rong Wen. 2022.
\newblock \href {https://aclanthology.org/2022.coling-1.552} {Context-tuning: Learning contextualized prompts for natural language generation}.
\newblock In \emph{International Conference on Computational Linguistics (COLING)}.

\bibitem[{Touvron et~al.(2023)Touvron, Martin, Stone, Albert, Almahairi, Babaei, Bashlykov, Batra, Bhargava, Bhosale et~al.}]{touvron-etal-2023-llama2}
Hugo Touvron, Louis Martin, Kevin Stone, Peter Albert, Amjad Almahairi, Yasmine Babaei, Nikolay Bashlykov, Soumya Batra, Prajjwal Bhargava, Shruti Bhosale, et~al. 2023.
\newblock \href {https://arxiv.org/pdf/2307.09288} {Llama 2: Open foundation and fine-tuned chat models}.
\newblock \emph{arXiv preprint arXiv:2307.09288}.

\bibitem[{Wadden et~al.(2019)Wadden, Wennberg, Luan, and Hajishirzi}]{wadden-etal-2019-entity}
David Wadden, Ulme Wennberg, Yi~Luan, and Hannaneh Hajishirzi. 2019.
\newblock \href {https://aclanthology.org/D19-1585} {Entity, relation, and event extraction with contextualized span representations}.
\newblock In \emph{Empirical Methods in Natural Language Processing and International Joint Conference on Natural Language Processing (EMNLP-IJCNLP)}.

\bibitem[{Wei et~al.(2022{\natexlab{a}})Wei, Bosma, Zhao, Guu, Yu, Lester, Du, Dai, and Le}]{wei-etal-2022-zeroshotlearners}
Jason Wei, Maarten Bosma, Vincent Zhao, Kelvin Guu, Adams~Wei Yu, Brian Lester, Nan Du, Andrew~M. Dai, and Quoc~V Le. 2022{\natexlab{a}}.
\newblock \href {https://openreview.net/forum?id=gEZrGCozdqR} {Finetuned language models are zero-shot learners}.
\newblock In \emph{International Conference on Learning Representations (ICLR)}.

\bibitem[{Wei et~al.(2022{\natexlab{b}})Wei, Wang, Schuurmans, Bosma, ichter, Xia, Chi, Le, and Zhou}]{wei-etal-2022-cot}
Jason Wei, Xuezhi Wang, Dale Schuurmans, Maarten Bosma, brian ichter, Fei Xia, Ed~Chi, Quoc~V Le, and Denny Zhou. 2022{\natexlab{b}}.
\newblock \href {https://proceedings.neurips.cc/paper_files/paper/2022/file/9d5609613524ecf4f15af0f7b31abca4-Paper-Conference.pdf} {Chain-of-thought prompting elicits reasoning in large language models}.
\newblock In \emph{Advances in Neural Information Processing Systems (NeurIPS)}.

\bibitem[{Yang et~al.(2024)Yang, Jin, Tang, Han, Feng, Jiang, Zhong, Yin, and Hu}]{yang2024harnessing}
Jingfeng Yang, Hongye Jin, Ruixiang Tang, Xiaotian Han, Qizhang Feng, Haoming Jiang, Shaochen Zhong, Bing Yin, and Xia Hu. 2024.
\newblock \href {https://doi.org/10.1145/3649506} {Harnessing the power of llms in practice: A survey on chatgpt and beyond}.
\newblock \emph{ACM Transactions on Knowledge Discovery from Data (TKDD)}.

\bibitem[{Yang et~al.(2023)Yang, Guo, Hu, Zhang, Qiu, and Zhang}]{yang-etal-2023-amr}
Yuqing Yang, Qipeng Guo, Xiangkun Hu, Yue Zhang, Xipeng Qiu, and Zheng Zhang. 2023.
\newblock \href {https://aclanthology.org/2023.acl-long.720} {An {AMR}-based link prediction approach for document-level event argument extraction}.
\newblock In \emph{Association for Computational Linguistics (ACL)}.

\bibitem[{Zeng et~al.(2023)Zeng, Liu, Du, Wang, Lai, Ding, Yang, Xu, Zheng, Xia, Tam, Ma, Xue, Zhai, Chen, Liu, Zhang, Dong, and Tang}]{zeng2023glmb}
Aohan Zeng, Xiao Liu, Zhengxiao Du, Zihan Wang, Hanyu Lai, Ming Ding, Zhuoyi Yang, Yifan Xu, Wendi Zheng, Xiao Xia, Weng~Lam Tam, Zixuan Ma, Yufei Xue, Jidong Zhai, Wenguang Chen, Zhiyuan Liu, Peng Zhang, Yuxiao Dong, and Jie Tang. 2023.
\newblock \href {https://openreview.net/forum?id=-Aw0rrrPUF} {{GLM}-130b: An open bilingual pre-trained model}.
\newblock In \emph{International Conference on Learning Representations (ICLR)}.

\bibitem[{Zeng et~al.(2022)Zeng, Zhan, and Ji}]{zeng-etal-2022-ea2e}
Qi~Zeng, Qiusi Zhan, and Heng Ji. 2022.
\newblock \href {https://aclanthology.org/2022.findings-naacl.202} {{EA}$^2${E}: Improving consistency with event awareness for document-level argument extraction}.
\newblock In \emph{Findings of North American Chapter of the Association for Computational Linguistics (NAACL)}.

\bibitem[{Zheng et~al.(2024)Zheng, Zhang, Zhang, Ye, and Luo}]{zheng2024llamafactory}
Yaowei Zheng, Richong Zhang, Junhao Zhang, Yanhan Ye, and Zheyan Luo. 2024.
\newblock \href {https://arxiv.org/pdf/2403.13372} {Llamafactory: Unified efficient fine-tuning of 100+ language models}.
\newblock \emph{arXiv preprint arXiv:2403.13372}.

\bibitem[{Zhou et~al.(2023)Zhou, Muresanu, Han, Paster, Pitis, Chan, and Ba}]{zhou-etal-2022-large}
Yongchao Zhou, Andrei~Ioan Muresanu, Ziwen Han, Keiran Paster, Silviu Pitis, Harris Chan, and Jimmy Ba. 2023.
\newblock \href {https://openreview.net/forum?id=92gvk82DE-} {Large language models are human-level prompt engineers}.
\newblock In \emph{International Conference on Learning Representations (ICLR)}.

\bibitem[{Zhu et~al.(2023)Zhu, Li, Liu, Ma, and Wang}]{zhu-etal-2023-survey}
Xunyu Zhu, Jian Li, Yong Liu, Can Ma, and Weiping Wang. 2023.
\newblock \href {https://arxiv.org/pdf/2308.07633} {A survey on model compression for large language models}.
\newblock \emph{arXiv preprint arXiv:2308.07633}.

\bibitem[{Zou et~al.(2021)Zou, Yin, Zhong, Yang, Yang, and Tang}]{zou-etal-2021-inverseprompting}
Xu~Zou, Da~Yin, Qingyang Zhong, Hongxia Yang, Zhilin Yang, and Jie Tang. 2021.
\newblock \href {https://doi.org/10.1145/3447548.3467418} {Controllable generation from pre-trained language models via inverse prompting}.
\newblock In \emph{International Conference on Knowledge Discovery and Data Mining (KDD)}.

\end{thebibliography}
\bibliographystyle{acl_natbib}

\newpage
\appendix
\section{Details of Inverse Prompting}\label{inverseprompting-appendix}
\subsection{Few-shot Instance of Inverse Prompting}
We provide a few-shot learning instance of manually labeled recovered context in Table~\ref{tab:recover-appendix}. We train the inverse prompting model using 97 question-rephrased context pairs generated by ChatGPT.
\begin{table}[h]
\centering
\begin{tabularx}{\linewidth}{X}
\toprule
\textbf{System:} \\
You are a helpful assistant. I'm giving you a question and an event trigger, please use them to recover the context of the event. Try your best to include as much information as possible. \\
\midrule
\textbf{User:} \\
trigger: bankruptcy question: What organization will declare bankruptcy soon? \\ 
\textbf{Assistant:}
An organization is soon to declare bankruptcy. \\
\midrule
\textbf{User:} \\
trigger: bankruptcy question: Where did WorldCom declare the bankruptcy? \\ 
\textbf{Assistant:}
WorldCom declared bankruptcy in somewhere. \\
\midrule
\textbf{User:} \\ 
trigger: fall question: What organization was ended by iraqis? \\ 
\textbf{Assistant:}
An organization was ended by Iraqis during a fall. \\
\midrule
\textbf{User:} \\ 
trigger: fallen question: Where did dictator Suharto fallen and democratic elections executed? \\ 
\textbf{Assistant:}
Dicatator Suharto was fallen and democratic elections were executed somewhere. \\
\midrule
\textbf{User:} \\ 
trigger: founded question: Who started the automaker in 1937? \\ 
\textbf{Assistant:}
Someone founded the automaker in 1937 \\
\bottomrule
\end{tabularx}
\caption{Few-shot learning instance of inverse prompting.}
\label{tab:recover-appendix}
\end{table}

\subsection{Variations of Inverse Prompting Model}
We utilize the supervised fine-tuning inverse prompting model in our setting. Moreover, we also discuss the performance with some variations of the inverse prompting model (IPM): 1) Removing the SFT process that directly uses an off-the-shelf model instead (specifically, LLaMA-2-7b-Chat). 2) Replacing the SFT process with 5-shot learning. The few-shot learning instances are collected from Table~\ref{tab:recover-appendix}.

\begin{table}[h]
\setlength\tabcolsep{5pt}
\centering
\begin{tabular}{lccc}
\hline
\textbf{Method} & \textbf{EM} & \textbf{COR} & \textbf{SemSim} \\ 
\hline
RLQG & 41.39 & 48.58 & 67.94  \\
-1) w/o SFT IPM & 40.35 & 47.37 & 66.12 \\
-2) w/ 5-shot IPM & 41.01 & 48.23 & 67.42 \\
\hline
\end{tabular}
\caption{Performance with Inverse Prompting Model Variations.}
\label{tab:ipm-ablation}
\end{table}
Table~\ref{tab:ipm-ablation} shows the corresponding results. Notice that directly recovering the context is challenging, but using few-shot examples obtains a solid performance without fine-tuning.

\section{Data Preprocessing}
\label{sec:appendix data-preprocessing}
\subsection{ACE2005}
We follow the step to preprocess and split the data in~\cite{wadden-etal-2019-entity}. To get the template question, the simple template~\cite{du-cardie-2020-event,liu-etal-2020-event} can be directly adopted, and the dynamic template~\cite{lu-etal-2023-event} needs to be filled in. We followed their proposed method to obtain the questions.

\subsection{RAMS}
The dataset is officially split and in natural language type in \href{https://nlp.jhu.edu/rams}{https://nlp.jhu.edu/rams}. We get the question by simply combining the opponent of trigger and role as equation~\ref{eq:prompt}. The back translation question is obtained by Google document translation\footnote{\href{https://translate.google.com/?sl=auto&tl=en&op=docs}{https://translate.google.com}} by translating the question to Chinese and back to English. 

\section{Implementation Details}\label{sec:appendix implementation}
We used parameter-efficient fine-tuning (PEFT) in the previous training stage to train our models. Specifically, in each stage (supervised fine-tuning and reinforcement learning), we utilize low-rank adaptation (LoRA)~\cite{hu-etal-2021-lora} as PEFT method, the trainable parameters occupy 0.0622\% of full parameters. Every random seed in our experiments is set to 42. The details of training and hyper-parameters are as follows.

\subsection{Supervised Fine-tuning}
As previously introduced, the base model of question generation we selected is LLaMA-2-7b; the training details are listed in Table~\ref{tab:sft-appendix}. The model's architecture is identical to the official provided in Huggingface. Also, we train the inverse prompting model with identical setting.
\begin{table}[h]
\centering
\setlength\tabcolsep{12pt}
\begin{tabular}{cc}
\hline
Hyper-parameters & Value \\
\hline
data type & bf16 \\
learning rate & 5e-05 \\
number of epochs & 3 \\
number of batch size & 16 \\ 
gradient accumulation steps & 4 \\
\hline
\end{tabular}
\caption{Hyper-parameters of the SFT stage of question generation model's training}
\label{tab:sft-appendix}
\end{table}\\
We tested our method's performance on different base model selections, including ChatGLM~\cite{zeng2023glmb} and Qwen~\cite{bai2023qwen}. The results are shown in Table~\ref{tab:selection-appendix}; our method also gets the best performance.

\begin{table}[h]
\centering
\setlength\tabcolsep{3pt}
\begin{tabular}{lccc}
\hline
\textbf{Models} & \textbf{EM} & \textbf{COR} & \textbf{SemSim} \\ 
\hline
ChatGLM-3-6b(SFT) & 38.63 & 45.09 & 64.67 \\
ChatGLM-3-6b(RLQG) & 40.45 & 46.97 & 66.16 \\
\midrule
Qwen-7b(SFT) & 41.45 & 48.48 & 67.43 \\
Qwen-7b(RLQG) & 43.22 & 50.04 & 69.08 \\
\hline
\end{tabular}
\caption{Performance on the ACE test set with different base model selections.}
\label{tab:selection-appendix}
\end{table}
We also verified our method on different model's size and variations, as shown in Table~\ref{tab:size-appendix}.

\begin{table}[h]
\centering
\setlength\tabcolsep{3pt}
\begin{tabular}{lccc}
\hline
\textbf{Model Size} & \textbf{EM} & \textbf{COR} & \textbf{SemSim} \\ 
\hline
LLaMA-2-7b & 41.39 & 48.58 & 67.94  \\
LLaMA-2-7b-Chat & 41.36 & 49.01 & 67.85 \\
LLaMA-2-13b & 41.53 & 48.77 & 67.81 \\
LLaMA-2-13b-Chat & 41.49 & 48.68 & 67.83 \\
\hline
\end{tabular}
\caption{Performance on the ACE test set with different model sizes and variations.}
\label{tab:size-appendix}
\end{table}

\subsection{PPO Training}
The hyper-parameters of reinforcement learning stage is similar with the previous stage, the details are shown in Table~\ref{tab:ppo-appendix}.
\begin{table}[h]
\setlength\tabcolsep{12pt}
\centering
\begin{tabular}{cc}
\hline
Hyper-parameters & Value \\
\hline
data Type & bf16 \\
learning rate & 1e-05 \\
number of epochs & 1 \\
number of batch size & 8 \\ 
gradient accumulation steps & 4 \\
\hline
\end{tabular}
\caption{Hyper-parameters of the RL stage of question generation model's training}
\label{tab:ppo-appendix}
\end{table}\\
Notice that the learning rate of reward model training is set to 1e-6.

\subsection{Generation Configuration}
In each generation part of our framework, including the training stage, question prediction, and question answering, the configurations are identical. The details are listed in Table~\ref{tab:gen-appendix}.
\begin{table}[h]
\setlength\tabcolsep{20pt}
\centering
\begin{tabular}{cc}
\hline
Configuration & Value \\
\hline
top p & 0.9 \\
do sample & True \\
temperature & 0.6 \\
max token length & 4096 \\
predict with generate & True \\
\hline
\end{tabular}
\caption{Generation configuration}
\label{tab:gen-appendix}
\end{table}\\
Notice that, in beam search augmentation in Section~\ref{sec:BSG}, the configurations are slightly different; the ``do sample'' option should be ``True'', and the number of beams is set to 10, and the number of return sentences is 5.

\subsection{Versions of Proprietary LLMs}
The version of ChatGPT in this work is GPT-3.5-Turbo-1106~\cite{openai-2023-gpt}, and version of GPT-4 is GPT-4-1106-preview~\cite{openai-2023-gpt}.

\section{Few-shot Learning Details}\label{sec:appendix fewshots}
In few-shot learning, the examples are directly combined with the input prompt by setting different characters in front of the context.

\subsection{Question Generation with Few-shot Learning}
In section~\ref{sec:exp setup}, we compared our method with method LLaMA-2-13b-Chat \textit{(5-shot)} and GPT-4 \textit{(5-shot)}. The prompt details are described in Table~\ref{tab:qgshots-appendix}.
\begin{table}[h]
\centering
\begin{tabularx}{\linewidth}{X}
\toprule
\textbf{System:} \\
You are a helpful assistant. Please generate a natural language question with contextual information. The question aims to help language models to extract roles from context. Generate a question based on the even trigger given. \\
\midrule
\textbf{User:} \\
role: agent trigger: election context: He lost an * election * to a dead man. \\ 
\textbf{Assistant:}
Who was the voting agent? \\
\midrule
\textbf{User:} \\
role: person trigger: quit context: Media tycoon Barry Diller on Wednesday * quit * as chief of Vivendi Universal Entertainment, the entertainment unit of French giant Vivendi Universal whose future appears up for grabs.\\ 
\textbf{Assistant:}
Who was quit as chief of Vivendi Universal Entertainment? \\
\midrule
\textbf{User:} \\
role: vehicle trigger: landed context: Even as the secretary of homeland security was putting his people on high alert last month, a 30-foot Cuban patrol boat with four heavily armed men * landed * on American shores, utterly undetected by the Coast Guard Secretary Ridge now leads. \\ 
\textbf{Assistant:}
What vehicle was used for transporting men to shores? \\
\midrule
\textbf{User:} \\
role: entity trigger: Former context: NOVAK * Former * Arkansas Governor and U.S. Senator Dale Bumpers has just published a memoir called "The Best Lawyer in a One-Lawyer Town." And it spans his life from the depression era to the Clinton era. \\ 
\textbf{Assistant:}
Who fired Dale Bumpers? \\
\midrule
\textbf{User:} \\
role: place trigger: war context: It could swell to as much as \$500 billion if we go to * war * in Iraq. \\ 
\textbf{Assistant:}
Where did the war take place? \\
\bottomrule
\end{tabularx}
\caption{Few-shot learning instance of question generation}
\label{tab:qgshots-appendix}
\end{table}

\subsection{Question Answering with Few-shot Learning}
The question answering in our research are conducted under calling a freeze chat model: LLaMA-2-13b-Chat with 5-shot. In the ablation study that using different models for question answering are still using the same prompt with identical examples. Details are listed in Table~\ref{tab:qashots-appendix}

\begin{table}[h]
\centering
\begin{tabularx}{\linewidth}{X}
\toprule
\textbf{System:} \\
You are a precise and concise assistant. Your task is to extract some words based directly on the provided context to answer the given questions. Please wrap your answer with the following tags: $[\text{ANS}]$ $[/\text{ANS}]$. If a question has multiple correct answers within the context, list them all, separated by commas. If there is no answer in the context, just reply $[\text{ANS}]$ None $[/\text{ANS}]$. Do NOT add any introductory phrases, explanations, or additional information outside of the given context. \\
\midrule
\textbf{User:} \\
question: Who made the battle in Baghdad? context: US Secretary of Defense Donald Rumsfeld dismissed worries that there were insufficient forces in the Gulf region if the battle for Baghdad goes wrong. \\ 
\textbf{Assistant:}
[ANS] US [/ANS] \\
\midrule
\textbf{User:} \\
question: Who was nominated? context: Senator Christopher Dodd of Connecticut made the announcement today that he would not be the 10th candidate for the nomination. \\ 
\textbf{Assistant:}
[ANS] candidate [/ANS] \\
\midrule
\textbf{User:} \\
question: Who is person in former event? context: We're talking about possibilities of full scale war with former Congressman Tom Andrews, Democrat of Maine. \\ 
\textbf{Assistant:}
[ANS] Tom Andrews [/ANS] \\
\midrule
\textbf{User:} \\
question: Who died that cause Clinton suffered greatly? context: Clinton suffered greatly over the 19 Rangers that died, 18 on the 3rd of October and Matt Reersen (ph) three days later. \\ 
\textbf{Assistant:}
[ANS] Rangers, Matt Reersen [/ANS] \\
\midrule
\textbf{User:} \\
question: Where did the election takes place? context: He lost an election to a dead man. \\ 
\textbf{Assistant:}
[ANS] None [/ANS] \\
\bottomrule
\end{tabularx}
\caption{Few-shot learning instance of question answering}
\label{tab:qashots-appendix}
\end{table}

\end{document}